\documentclass[letterpaper, 10 pt, conference]{ieeeconf}  % Comment this line out if you need a4paper

\IEEEoverridecommandlockouts                              % This command is only needed if 
\title{\LARGE \bf
Mind the Gap: Evaluating Patch Embeddings from General-Purpose and Histopathology Foundation Models for Cell Segmentation and Classification
}

\author{Valentina Vadori$^{1}$, Antonella Peruffo$^{2}$, Jean-Marie Graïc$^{2}$, Livio Finos$^{3}$, and Enrico Grisan$^{1}$% <-this % stops a space
\thanks{$^{1}$Dept. of Computer Science and Informatics,
        London South Bank University, London, UK,
        {\tt\small vadoriv@lsbu.ac.uk}}%
\thanks{$^{2}$Dept. of Comparative Biomedicine and Food Science, University of Padova,
        Padova, Italy}%
        \thanks{$^{3}$Dept. of Statistical Sciences, University of Padova,
        Padova, Italy}%
}

\usepackage{cite}

\usepackage{graphicx}
\usepackage{textcomp}
\usepackage{url}
\usepackage{booktabs}
\usepackage[table,xcdraw,dvipsnames]{xcolor}
\usepackage{soul} % Better underline control
\usepackage{colortbl}
\usepackage{nicematrix}
\usepackage{caption}
\usepackage{amsmath,amssymb,amsfonts}
\usepackage{algorithmic}
\definecolor{custompurple}{rgb}{0.55, 0.0, 1}
\definecolor{customteal}{HTML}{0.C0C1}
\definecolor{lightgrey}{HTML}{bfbfbf}
\setulcolor{lightgrey}
\setul{0.5pt}{0.4pt}% 1pt below contents
\begin{document}

\maketitle
\thispagestyle{empty}
\pagestyle{empty}

%%%%%%%%%%%%%%%%%%%%%%%%%%%%%%%%%%%%%%%%%%%%%%%%%%%%%%%%%%%%%%%%%%%%
\begin{abstract}
Recent advancements in foundation models have transformed computer vision, driving significant performance improvements across diverse domains, including digital histopathology. However, the advantages of domain-specific histopathology foundation models over general-purpose models for specialized tasks such as cell analysis remain underexplored.  This study investigates the representation learning gap between these two categories by analyzing multi-level patch embeddings applied to cell instance segmentation and classification. We implement an encoder-decoder architecture with a consistent decoder and various encoders. These include convolutional, vision transformer (ViT), and hybrid encoders pre-trained on ImageNet-22K or LVD-142M, representing general-purpose foundation models. These are compared against ViT encoders from the recently released UNI, Virchow2, and Prov-GigaPath foundation models, trained on patches extracted from hundreds of thousands of histopathology whole-slide images. The decoder integrates patch embeddings from different encoder depths via skip connections to generate semantic and distance maps. These maps are then post-processed to create instance segmentation masks—where each label corresponds to an individual cell—and to perform cell-type classification. All encoders remain frozen during training to assess their pre-trained feature extraction capabilities. Using the PanNuke and CoNIC histopathology datasets, and the newly introduced Nissl-stained CytoDArk0 dataset for brain cytoarchitecture studies, we evaluate instance-level detection, segmentation, and cell-type classification accuracy. This study provides insights into the comparative performance of general-purpose vs. histopathology foundation models, offering guidance for model selection in cell-focused histopathology and brain cytoarchitecture analysis workflows.
%\newline
%\indent \textit{Clinical relevance}— This is a brief additional statement on why a this might be of interest to practicing clinicians. Example: This establishes the anesthetic efficacy of 10\% intraosseous injections with epinephrine to positively influence cardiovascular function.
\end{abstract}

%%%%%%%%%%%%%%%%%%%%%%%%%%%%%%%%%%%%%%%%%%%%%%%%%%%%%%%%%%%%%%%%%%%%%%%%%%%%%%%%
\section{INTRODUCTION}

In recent years, foundation models have emerged as powerful tools in computer vision, driving significant performance improvements across various domains, including computational histopathology. Large-scale deep neural networks, tipically vision transformers (ViTs), are pre-trained on extensive histopathology data encompassing diverse stains, tissue types, and diseases, typically without task-specific labels. By applying self-supervised learning approaches, these models learn image representations that generalize effectively across a range of downstream tasks, such as diagnosis, disease subtyping, biomarker quantification, treatment response estimation, and survival prediction \cite{wang2024pathology, lu2024visual, xu2024whole, chen2024uni, vorontsov2024foundation, zimmermann2024virchow2}. 
\begin{figure*}
\centering
\includegraphics[width=0.95\linewidth]{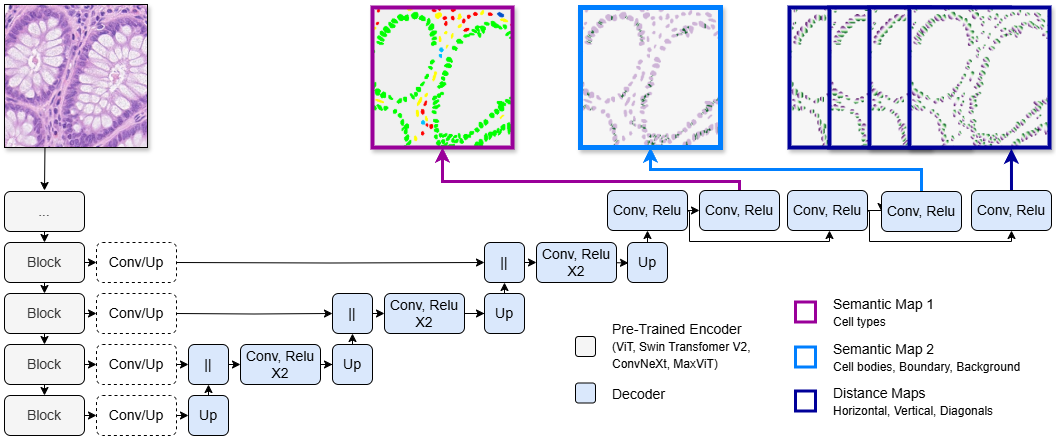}
\caption{A schematic representation of the encoder-decoder architecture utilized in this study. Feature maps are extracted from four encoder blocks and processed through convolution and upsampling operations to achieve predefined channel dimensions and resolutions. The decoder, consisting of upsampling and convolutional blocks, is designed to generate four distinct outputs based on the CISCA framework \cite{vadori2024cisca}. These outputs are further post-processed to produce the final label map and to assign a specific cell type to each detected cell.} \label{Architecture}
\end{figure*}
Recent studies demonstrate the performance gains of \textit{domain-specific} (in the following, also \textit{specialized}) foundation models but their advantages over \textit{general-purpose} models, such as CNNs or hybrid architectures trained on the ImageNet dataset, remain underexplored for key histopathology tasks, including cell instance segmentation (CS) and cell classification (CC). Traditional frameworks for CS and CC, such as Hover-Net \cite{graham2019hover}, StarDist \cite{schmidt2018cell}, and Cellpose \cite{stringer2021cellpose}, are widely used due to their efficiency and adaptability to domain-specific data. However, the extent to which foundation models can be integrated into these frameworks and leveraged to surpass state-of-the-art pipelines for CS and CC requires further investigation. CellViT \cite{horst2024cellvit}, a UNETR architecture \cite{hatamizadeh2022unetr} that leverages the Hover-Net framework and exploits a pre-trained foundation model as the encoder,
% similar to the one considered in this study with a pre-trained foundation model as the encoder, 
has recently demonstrated superior performance over state-of-the-art approaches; however, its evaluation was limited to the PanNuke dataset, and it was not compared against similar architectures using foundation models as encoders. CONCH \cite{lu2024visual} has not been applied to cell or nuclei segmentation, making its suitability for such tasks unclear. The authors of UNI \cite{chen2024uni}  considered a ResNet trained on ImageNet as a general-purpose baseline, despite prior studies indicating that ConvNeXt \cite{liu2022convnet} and Swin Transformer \cite{liu2021swin} architectures typically outperform ResNet for vision tasks. Virchow2 \cite{zimmermann2024virchow2} has been tested on cell detection but its performance on CS and CC has not yet been evaluated.

This study aims to quantify performance differences between histopathology-specific  and general-purpose foundation models for CS and CC. We evaluate the recently introduced UNI2, Virchow2, and Prov-GigaPath \cite{xu2024whole} foundation models, comparing them against convolutional, ViT, and hybrid networks pre-trained on ImageNet-22K or LVD-142M. Our evaluation is conducted on three histology datasets: PanNuke and CoNIC H\&E histopathology datasets, with 40x and 20x magnification, respectively, and the newly introduced CytoDArk0 dataset, which consists of Nissl-stained histology slices of brain cells from various mammalian species \cite{vadori2024cytodark0}. This selection allows us to assess model generalization across different cell types and staining methods.

By analyzing key performance metrics, including instance-level detection, segmentation accuracy, and cell-type classification, this study reveals a performance gap that should be addressed—though not in the expected direction. Our findings offer practical guidance for model selection and development in cell-focused histopathology and brain cytoarchitecture analysis workflows. %Ultimately, this work contributes to the optimization of automated analysis pipelines, enabling more accurate and efficient biomedical research and clinical applications.

\section{METHODS}
We utilize the CISCA framework \cite{vadori2024cisca} for CS and CC, which has demonstrated greater robustness and accuracy compared to state-of-the-art methods such as Hover-Net and StarDist across different staining techniques, magnifications, and tissue types. CISCA addresses CS and CC using a multi-task approach that integrates 3-class pixel classification, distance map regression, and cell-type classification. Specifically, the framework predicts a semantic map (SM1) that assigns each pixel to one of three categories: cell body, boundary between neighboring cells, or background. Additionally, it generates four directional distance maps (DMs), and a second semantic map (SM2) where each pixel is classified by cell type. Post-processing is then applied to produce a label map, where each unique cell is assigned an integer ID, and to define the type of each detected cell.
To implement CISCA, we adopt an encoder-decoder architecture similar to the original design, as illustrated in Fig. \ref{Architecture}. In line with \cite{vadori2024cisca}, we employ one decoder with three convolutional heads to predict SM1, SM2 and DMs. However%to accommodate different encoder architectures
, as described in Section \ref{Decoder}, the decoder and skip connections design is inspired by the UNETR architecture  \cite{hatamizadeh2022unetr} rather than the original U-Net. This design choice ensures compatibility with the encoder-decoder architecture implementation in the \texttt{cellseg\_models.pytorch} \cite{csmp2022} library, facilitating the adoption of a variety of encoders from the \texttt{timm} library, including ViTs. %. and the potential testing of other paradigms such as Cellpose, Hover-Net, and Stardist in the future.
Following the CISCA approach, we apply the same oversampling technique for the PanNuke and CoNIC datasets and use an identical data augmentation strategy. No pre-processing is performed on the input patches. For further details on CISCA, we refer to \cite{vadori2024cisca}. 
\subsection{Encoders}
\label{Encoders}
We utilize four state-of-the-art architectures as encoders, which are either pre-trained on general-purpose datasets, i.e., the publicly available ImageNet-22K dataset \cite{deng2009imagenet} (a superset of the ImageNet-1K dataset with $\sim$ 14M images), ImageNet-21K (Google specific variant of ImageNet-22K), or the proprietary LVD-142M dataset \cite{oquab2023dinov2}, or on specialized datasets, i.e., extensive, private histopathology datasets. %While models pre-trained on ImageNet-22K are not classified as foundation models, the others are.  
\begin{itemize}
\item \textbf{ViT} \cite{alexey2020image}: ViT adapts the transformer architecture, originally developed for natural language processing \cite{NIPS2017_3f5ee243}, to image processing by partitioning images into sequences of flattened 2D \textit{patches}, which are linearly projected into embeddings. These embeddings, combined with positional encodings to retain spatial context, are processed through multiple layers of multi-head self-attention and feedforward networks. This design enables ViT to capture long-range dependencies in visual data, outperforming ResNet-based baselines in image classification tasks.
\item \textbf{Swin Transformer} \cite{liu2021swin}: In contrast to the quadratic complexity of ViTs, the Swin Transformer achieves linear computational complexity by limiting self-attention computations to non-overlapping local \textit{windows}, thereby reintroducing the locality bias inherent in convolutional networks. Additionally, it constructs a hierarchical representation by progressively merging adjacent patches in deeper layers, facilitating seamless integration with dense prediction frameworks such as U-Net. In this study, we focus on Swin Transformer V2 \cite{liu2022swin}, which enhances the scalability of its predecessor with multiple adjustments.
%This design makes the Swin Transformer particularly effective for high-resolution vision tasks.
%Cross-window interactions are achieved by shifting these windows across successive blocks. Furthermore, it builds a hierarchical representation by initially segmenting images into small patches and gradually merging adjacent patches in deeper layers.
\item \textbf{ConvNeXt} \cite{liu2022convnet}: A modernized ResNet architecture that incorporates design elements from the Swin Transformer (e.g., similar training recipe, $1$:$1$:$3$:$1$ or $1$:$1$:$9$:$1$ stage compute ratio, $4\times4$, stride $4$ convolutional layer as a ``patchify stem'', depthwise convolution, more feature channels, larger kernel sizes, fewer activation functions) achieving improved performance in image classification, detection, and segmentation tasks with comparable FLOPs, without using attention modules.
\item \textbf{MaxViT} \cite{tu2022maxvit}: To overcome the non-locality limitations of the Swin Transformer and the high computational cost of full-attention in ViTs, MaxViT introduces a hybrid architecture that alternates between convolutional layers and Max-SA, a multi-axis attention mechanism. Max-SA consists of a block-attention module for local self-attention within fixed-size windows and a grid-attention module for capturing global dependencies, both operating with linear complexity relative to input size.
%Max-SA comprises a block-attention module, which performs local self-attention within fixed-size windows, followed by a grid-attention module that captures global dependencies by attending to patches in a sparse, uniform grid across the entire spatial domain. Both attention mechanisms achieve linear complexity with respect to input size.
%, ensuring scalability for high-resolution visual tasks.
\end{itemize}
We evaluate a Swin Transformer V2, a ConvNeXt, a MaxViT, and a ViT pre-trained on ImageNet-22K or ImageNet-21K using supervised learning, as well as a ViT pre-trained on LVD-142M using unsupervised learning. As specialized models, we consider the following.
\begin{itemize}
 %UNI is trained on a dataset of over 100 million images derived from more than 100,000 diagnostic H\&E-stained WSIs spanning 20 major tissue types. 
\item \textbf{UNI2}: An enhanced version of UNI \cite{chen2024uni}, UNI2 has been trained on a dataset of over  200 million $256\times256$ tissue patches extracted from more than 350,000   hematoxylin and eosin (H\&E) and immunohistochemistry (IHC)-stained WSIs at $20\times$ magnification, covering at least 20 tissue type (according to \cite{chen2024uni}). The WSIs are sourced from Mass General Brigham. The model weights for UNI2 were publicly released on January 14, 2025.
\item \textbf{Virchow2} \cite{zimmermann2024virchow2}: This model has been trained on $392\times392$ patches from 3.1 million H\&E-stained and IHC WSIs across multiple magnifications ($5\times$, $10\times$, $20\times$, and $40\times$), covering nearly 200 tissue types. Samples are collected from biopsies and resections of 225,401 patients of the Memorial Sloan Kettering Cancer Center and various other international institutions.
%collected from 225,401 patients.
%In addition to the samples provided by the Memorial Sloan Kettering Cancer Center (MSKCC), 15\% of the WSIs and 57\% of the patient data were obtained from various international institutions that submitted challenging cases to MSKCC for consultation.
\item \textbf{Prov-GigaPath} \cite{xu2024whole}: Prov-GigaPath is trained on nearly 1.4 billion $256\times256$ patches extracted from 171,189 H\&E-stained and IHC WSIs at $20\times$ magnification, covering 31 tissue types. The WSIs originate from biopsies and resections of more than 30,000 patients of 28 cancer centers of the Providence health network. 
\end{itemize}
The ViT architecture underpins these three foundation models, which have been pre-trained using either standard or customized versions of the self-supervised learning paradigm DINOv2 \cite{oquab2023dinov2}. They differ in the number of blocks, input image sizes, and patch sizes (see Section \ref{results}, Table \ref{pannukeperformance} for details).

%UNI2, Virchow2, and Prov-GigaPath utilized a "huge" ViT model (ViT-H) with a patch size of 14 and an input image size of 224×224. Consequently, images from the datasets were cropped accordingly to fit the model's input requirements during training.
\subsection{Decoder}
\label{Decoder}
The decoder and skip connections design is inspired by the UNETR architecture \cite{hatamizadeh2022unetr}. Feature maps from the encoder are extracted via skip connections at four distinct levels and fused with upsampled feature maps from the preceding, deeper layer. This fusion involves concatenation, $3\times3$ convolution, ReLU activation. At the top of the decoder, convolutional heads are employed to mitigate ``fighting for features'' issues during semantic maps and distance maps predictions. In cases where feature maps from the encoder are not hierarchical—such as in the ViT architecture, where the resolution remains consistent across all levels—feature maps are first processed through a $1\times1$ convolutional block to align with a predefined number of feature channels and then upsampled to a specified resolution. For the four levels, we define channel sizes as (32, 64, 128, 256) with resolution reductions of (2, 4, 8, 16). For example, given an input image of size $256\times256$, if the feature maps from the second and third level both have dimensions (8, 1280, 16, 16), they are processed to produce feature maps of size (8, 64, 64, 64) and (8, 128, 32, 32), respectively. 
%—where the width and height are 16 times smaller than the input image, and 8 represents the batch size—

%\newcommand{\grayuline}[1]{{\color{gray}\uline{{\color{black}#1}}}}

\subsection{Training}
\setlength{\tabcolsep}{3pt}
\begin{table*}[]
%\captionsetup{aboveskip=3pt, belowskip=0pt}
\caption{Cell instance segmentation and classification performance on the PanNuke dataset}
\label{pannukeperformance}
\centering
\begin{tabular}{@{}ccccc
>{\columncolor[HTML]{EFEFEF}}c cccccccc@{}}
\toprule
\textbf{Model} & \textbf{Type} & \textbf{Resolution} & \textbf{Patch size} & \textbf{\#blocks} & \textbf{Feature block} & \textbf{P}     & \textbf{R}     & \textbf{DQ (F1)} & \textbf{SQ}    & \textbf{PQ}    & \textbf{mPQ+}  & \textbf{\#P. (G)} & \textbf{FLOPs (G)} \\ \midrule
\arrayrulecolor{lightgray}
ConvNeXt-B-22K & ConvNeXt      & 224x224             & -                   & 36                & {[}2,5,32,35{]}        & 76.20          & 76.89          & 75.72            & \textbf{81.12} & \ul{62.27}    & 47.63          & \ul{0.115}       & \textbf{27}        \\
MaxViT-B-21K   & MaxViT        & 224x224             & -                   & 24                & {[}1,7,21,23{]}        & 75.71          & \textbf{77.71} & 75.85            & 80.81          & 62.23          & \ul{48.05}    & 0.238             & 64                 \\
Swin2-B-22K    & SwinT V2      & 192x192             & -                   & 24                & {[}1,3,21,23{]}        & \textbf{78.15} & 76.93          & \textbf{76.80}   & \ul{80.92}    & \textbf{63.19} & \textbf{50.32} & \textbf{0.114}    & \ul{29}           \\ \midrule
ViT-L-21K      & ViT           & 224x224             & 16x16               & 24                & {[}2,4,6,8{]}          & 74.58          & 73.54          & 73.34            & 77.10          & 57.79          & 44.19          & 0.308             & 41                 \\
ViT-H-142M    & ViT           & 518x518             & 14x14               & 40                & {[}2,4,6,8{]}          & 77.06          & 75.29          & 75.31            & 80.10          & 61.22          & 45.97          & 1.141             & 78                 \\
Prov-GigaPath & ViT           & 224x224             & 14x14               & 40                & {[}2,4,6,8{]}          & 75.15          & 74.69          & 73.99            & 79.52          & 59.63          & 42.65          & 1.141             & 78                 \\
UNI2         & ViT           & 224x224             & 14x14               & 24                & {[}2,4,6,8{]}          & 77.82          & 76.15          & 76.28            & 80.18          & 62.18          & 47.56          & 0.687             & 80                 \\
Virchow2     & ViT           & 224x224             & 14x14               & 32                & {[}2,4,6,8{]}          & 76.58          & 76.57          & 75.84            & 79.78          & 61.53          & 46.97          & 0.637             & 58                 \\ \midrule
ViT-L-22K      & ViT           & 224x224             & 16x16               & 24                & {[}17, 19, 21, 23{]}   & 69.70          & 66.99          & 67.50            & 73.59          & 51.12          & 37.43          & 0.308             & 89                 \\
ViT-H-142M    & ViT           & 518x518             & 14x14               & 40                & {[}34, 36, 37, 39{]}   & 65.79          & 62.80          & 63.41            & 70.89          & 46.52          & 31.27          & 1.141             & 303                \\
Prov-GigaPath & ViT           & 224x224             & 14x14               & 40                & {[}34, 36, 37, 39{]}   & 72.82          & 70.71          & 71.02            & 74.35          & 54.02          & 39.69          & 1.141             & 303                \\
UNI2         & ViT           & 224x224             & 14x14               & 24                & {[}17, 19, 21, 23{]}   & 72.92          & 70.02          & 70.66            & 73.77          & 53.46          & 38.67          & 0.687             & 192                \\
Virchow2     & ViT           & 224x224             & 14x14               & 32                & {[}25, 27, 29, 31{]}   & 74.70          & 71.70          & 72.35            & 75.17          & 55.51          & 41.04          & 0.637             & 176                \\ \midrule
ViT-L-22K      & ViT           & 224x224             & 16x16               & 24                & {[}2, 5, 10, 20{]}     & 72.20          & 70.98          & 70.80            & 75.61          & 54.71          & 40.75          & 0.308             & 80                 \\
ViT-H-142M    & ViT           & 518x518             & 14x14               & 40                & {[}2, 5, 18, 36{]}     & 75.64          & 73.00          & 73.44            & 76.76          & 57.38          & 41.98          & 1.141             & 281                \\
Prov-GigaPath & ViT           & 224x224             & 14x14               & 40                & {[}2, 5, 18, 36{]}     & 77.33          & 75.66          & 75.78            & 78.28          & 60.23          & 46.62          & 1.141             & 281                \\
UNI2         & ViT           & 224x224             & 14x14               & 24                & {[}2, 5, 10, 20{]}     & 77.33          & \ul{77.42}    & \ul{76.59}      & 78.81          & 61.39          & 47.37          & 0.687             & 170                \\
\arrayrulecolor{black}
Virchow2     & ViT           & 224x224             & 14x14               & 32                & {[}2, 5, 14, 28{]}     &\ul{78.11}    & 76.21          & 76.34            & 78.89          & 61.16          & 47.68          & 0.637             & 161                \\ \bottomrule
%UNI2          & ViT           & 224x224             & 14x14               & 24                & {[}1,3,21,23{]} &                & \ul{}         & \ul{}           &                &                &                &                   0.687             & 170                    \\
%\arrayrulecolor{black}
%Virchow2      & ViT           & 224x224             & 14x14               & 32                & {[}1,3,21,23{]} & \ul{ }         &                &                  &                &                &                &                   0.637             & 161                       \\ \bottomrule
\end{tabular}
\end{table*}

The training process aims to minimize a composite loss function that includes categorical cross-entropy loss and Dice loss for 3-pixel class prediction, mean absolute error (MAE) for distance map regression, and a combination of categorical cross-entropy loss and Tversky loss for cell type classification. Encoders are kept frozen during training. The decoder and skip connections are optimized using the AdamW optimizer and the OneCycleLR learning rate scheduler, which anneals the learning rate to a peak value of $10^{-4}$ and then to $10^{-6}$. The model is trained for a total of $100$ epochs, with performance evaluated after each epoch. %using the Jaccard Index and MAE on the validation set. 
The best-performing model on the validation set is retained for evaluation on the test set. 
\section{EXPERIMENTS}
\subsection{Models}
We evaluated eight different models, each named after the encoder it utilizes: ConvNeXt-B-22K, MaxViT-B-22K, Swin2-B-22K, ViT-L-22K , ViT-H-142M, UNI2, Prov-GigaPath, Virchow2\footnote{Pre-trained encoders available on Hugging Face: convnext\_base.fb\_in22k, maxvit\_large\_tf\_224.in21k, swinv2\_base\_window12\_192.ms\_in22k, vit\_large\_patch16\_224.orig\_in21k, vit\_giant\_patch14\_dinov2.lvd142m, prov-gigapath, UNI2-h, Virchow2.}. To evaluate whether feature maps from ViT blocks are sufficient for tasks such as CS and CC—and whether they retain localization information—we experiment with different strategies for feature extraction. Specifically, we consider three configurations for feature maps extraction: \textit{shallow}, extraction from the first eight ViT blocks; \textit{deep}, extraction from the last eight blocks; and \textit{mixed}, extraction across all blocks of the ViT architecture. For instance, in a ViT model with 32 blocks indexed from 0 to 31, we extract feature maps from blocks (2, 4, 6, 8) for the shallow configuration, (25, 27, 29, 31) for the deep configuration, and (2, 5, 14, 28) for the mixed configuration. 
This results in 18 distinct `models', which we evaluated on the PanNuke dataset. The top five models, defined as those ranking first or second in performance for at least one metric, were then chosen for further training and testing on the CoNIC and CytoDArk0 datasets. A description of the datasets is provided in the following.
\subsection{Datasets}
\begin{itemize}
\item \textbf{PanNuke} \cite{gamper2020pannuke} The largest pan-cancer nuclear instance segmentation and classification dataset, with nearly $200,000$ labelled nuclei in  $7,901$ H\&E-stained histology images of size $256\times256$ from $19$ different tissue types at $40$x. Each nucleus is assigned to one of $5$ classes: Neoplastic, Epithelial, Connective/soft tissue cells, Inflammatory, Dead cells. 

%\item 
\item \textbf{CoNIC Challenge Dataset} \cite{graham2021lizard, graham2024conic} The largest single tissue nuclear instance segmentation and classification dataset, with nearly $500,000$ labelled nuclei in  $4,981$ H\&E-stained colon tissue histology images of size $256\times256$ at $20$x. Each nucleus is assigned to one of $6$ classes: Neutrophil, Epithelial, Lymphocyte, Plasma, Eosinophil, Connective tissue cells.  

\item \textbf{CytoDArk0} \cite{vadori2024cytodark0} The first publicly available annotated dataset of Nissl-stained histological images of the mammalian brain for CS, comprising a total of $38,755$ cells, including neurons and glial cells. We use CytoDArk0\_20x\_256, a patched version of CytoDArk0 that consists of $1,104$ patches at $20$x, each with a size of $256$×$256$ pixels. CytoDArk0 was recently released to facilitate research in digital neuropathology and studies on brain cytoarchitecture \cite{graic2023cytoarchitectureal, graic2024age}.
\end{itemize}
%For ViTs, we explored three different strategies for extracting feature maps from the encoder: \textit{superficial} (four levels selected from the first eight), \textit{deep} (four levels from the last eight layers), and \textit{distributed} (four levels spanning the entire depth of the encoder). 

Each dataset used for training was split into three subsets, following \cite{vadori2024cisca}: approximately 70\% for training, 10\% for validation, and 20\% for testing. Details for replicating these splits can be found at \url{https://github.com/vadori/cytoark}. 
\subsection{Performance Evaluation}
CS performance is assessed by calculating the \textit{panoptic quality} ($PQ$) \cite{graham2024conic}. $PQ$ is the product of detection quality ($DQ$), which evaluates the model's ability to identify cell locations, and segmentation quality ($SQ$), which determines the accuracy of delineating cell boundaries. A predicted cell instance is a true positive ($TP$) if its intersection over union ($IoU$) with the corresponding ground truth instance is greater than 0.5. Predicted instances without a corresponding match are classified as false positives ($FP$), while unmatched ground truth instances are counted as false negatives ($FN$). We provide results for recall $R=|TP|/(|TP|+|FN|)$ and precision $P= |TP|/(|TP|+|FP|)$. The $F1$-score  corresponds to $DQ=|TP|/(|TP|+0.5\cdot(|FP|+|FN|))$, whereas $SQ$ is computed as the average $IoU$ between matched predicted instances 
%$\hat{s}$ 
and their respective ground truth instances. All the above metrics are computed separately for each
image, and the results are averaged. CC performance is assessed using the \textit{multi-class panoptic quality} \textit{mPQ+} \cite{graham2024conic}. This entails computing panoptic quality for each cell type by aggregating $TP$, $FP$, and $FN$ across all images, ensuring robustness to missing cell classes. The final metric is then averaged across classes.
%, applying the same formula as in Eq. \ref{eq:panoptic}. 

\setlength{\tabcolsep}{2pt}
\begin{table}[]
\centering
%\captionsetup{aboveskip=3pt, belowskip=0pt}
\caption{Performance on the CoNIC dataset}
\label{conicperformance}
\begin{tabular}{@{}c
>{\columncolor[HTML]{EFEFEF}}c cccccc@{}}
\toprule
\textbf{Model} & \textbf{Feature block} & \textbf{P} & \textbf{R} & \textbf{DQ} & \textbf{SQ} & \textbf{PQ} & \textbf{mPQ+} \\ \midrule
ConvNeXt-B-22K & {[}2,5,32,35{]}        &   \textbf{78.07}   & \textbf{77.08}  &  \textbf{77.27}                 &  \textbf{78.85}            &  \textbf{61.15}   &  \ul{43.79}              \\
MaxViT-B-21K   & {[}1,7,21,23{]}        &  76.55           &  \ul{76.87}          &  \ul{76.35}                & 77.83            &  59.67            &  \textbf{44.04}              \\
\arrayrulecolor{lightgray}
Swin2-B-22K    & {[}1,3,21,23{]}        &  \ul{77.53}           &  75.18           &  76.02                 &  \ul{78.59}            &  \ul{60.02}            &  43.50              \\ \midrule
UNI2          & {[}2, 5, 10, 20{]}     &  74.21           &  69.70           & 71.55                &  73.65            &  52.94            &  40.47              \\
\arrayrulecolor{black} 
Virchow2      & {[}2, 5, 14, 28{]}     &  72.98           &  69.94           &  71.06                 &  73.66            &  52.59            &  39.64              \\ 
\bottomrule
\end{tabular}
\end{table}
\begin{table}[]
\centering
%\captionsetup{aboveskip=3pt, belowskip=0pt}
\caption{Performance on the CytoDArk0 dataset}
\label{cytodark0performance}
\begin{tabular}{@{}c
>{\columncolor[HTML]{EFEFEF}}c ccccc@{}}
\toprule
\textbf{Model} & \textbf{Feature block} & \textbf{P} & \textbf{R} & \textbf{DQ} & \textbf{SQ} & \textbf{PQ}  \\ \midrule
ConvNeXt-B-22K & {[}2,5,32,35{]}        & \ul{76.18}            & \textbf{81.00}           & \textbf{77.67}                & \ul{83.68}            & \textbf{65.07}                      \\
MaxViT-B-21K   & {[}1,7,21,23{]}        &  48.52          &  \ul{79.31}            &  58.38              &  79.03            &  46.65                         \\
\arrayrulecolor{lightgray}
Swin2-B-22K    & {[}1,3,21,23{]}        &  \textbf{76.42}        &  79.09       &  \ul{77.04}             &  \textbf{83.70}        &  \ul{64.59}   \\ \midrule
UNI2          & {[}2, 5, 10, 20{]}     &  67.73           &  71.43           &  68.19                 &  76.67            &  52.85                        \\
\arrayrulecolor{black} 
Virchow2      & {[}2, 5, 14, 28{]}     &  70.89           &  72.29           &  70.50                 &  76.98           &  54.83                         \\\bottomrule

\end{tabular}
\end{table}
\subsection{Implementation}
Our implementation is built on the PyTorch framework, incorporating custom code to the source code of the Python library \texttt{cellseg\_models.pytorch} \cite{csmp2022}, 
which leverages the \texttt{timm} library to import pre-trained encoders \cite{rw2019timm}. The training was carried out on an NVIDIA Quadro RTX 6000 and an NVIDIA RTX A5000, each with 24 GB of memory, on a system running Windows 10 with 128 GB of RAM. Source code will be released at \url{https://github.com/vadori/cytoark}.

\begin{figure}
\centering
\includegraphics[width=\linewidth]{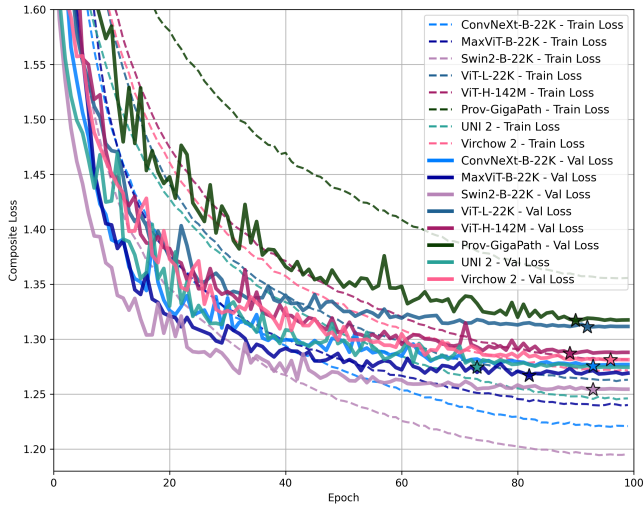}
%\captionsetup{aboveskip=3pt, belowskip=0pt}
\caption{Loss curves over 100 epochs for different models trained on PanNuke. Dotted lines and and solid lines represent training and validation loss, respectively. The lowest validation loss for each model (whose name is inherited from the encoder used) is marked for reference.} \label{losscurves1}
\end{figure}

\section{RESULTS}
\label{results}

CS and CC performance on the PanNuke dataset for each model evaluated in this study are shown in Table \ref{pannukeperformance}. The horizontal lines categorize different models according to the encoder types: non-ViT models (top section); shallow ViT configurations (second section); deep ViT configurations (third section); mixed configurations (bottom section), depending on the ViT blocks used for feature extraction. 

The Swin2-B-22K model emerges as the top-performing method overall, achieving the highest scores across multiple key metrics. It excels in PQ, demonstrating its ability to accurately detect and segment cell contours, and in mPQ+, reflecting its effectiveness in classifying cells into distinct types. Additionally, Swin2-B-22K maintains computational efficiency, requiring the fewest parameters and the second-lowest FLOPs among all models. MaxViT-B-22K achieves the best result on R and the second-best result on mPQ+. ConvNeXt-B-22K achieves the best result on SQ and the second-best result on PQ. ViT models are outperformed by non-ViT models across all indicators with a few exceptions. Virchow2 (mixed) ranks second in precision, while UNI2 (mixed) achieves the second-highest recall and DQ. 

For specialized foundation models, shallow ViT configurations outperform deeper ViT configurations that rely on features from deeper blocks. Mixed configurations enhance performance in DQ and mPQ+ for ProvGiga-Path and Virchow2, though the improvement is moderate. For UNI2 the shallow variant remains the strongest across these metrics. This is also true for general-purpose ViTs, whose performance decline in deep and mixed configurations. On the contrary, ProvGiga-path falls behind in the shallow configuration but competes well with other specialized foundation models in the deep and mixed configurations.
Overall, general-purpose ViTs tend to be outperformed by specialized ViTs.
It can be noted that shallow configurations excel in SQ, as local and fine-grained information is crucial for high-quality segmentation and is plausibly lost in deeper blocks.

%However, Swin2-B-22K outperforms all models, across indicators, except for R and SQ, where 

\begin{figure}
\centering
\includegraphics[width=\linewidth]{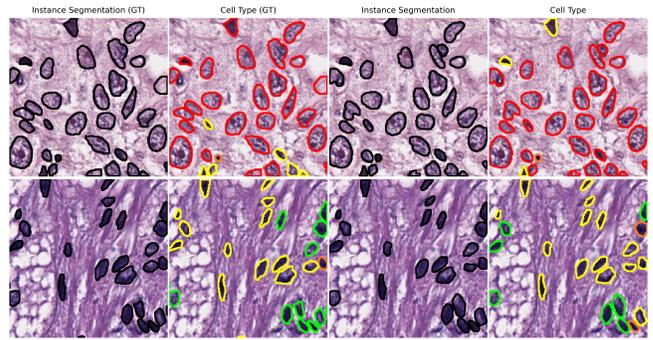}
%\captionsetup{aboveskip=2pt, belowskip=0pt}
\caption{Example of cell instance segmentation and classification on two test patches from PanNuke using Swin2-B-22K.}
\label{pannukequalit}
\end{figure}
\begin{figure}
\centering
\includegraphics[width=\linewidth]{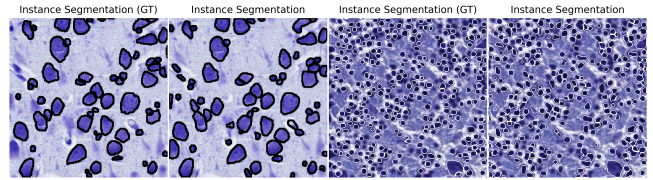}
%\captionsetup{aboveskip=2pt, belowskip=0pt}
\caption{Example of cell instance segmentation on two test patches from CytoDArk0 using ConvNeXt-B-22K.} \label{cytodark0equalit}
\end{figure}
Figure \ref{losscurves1} illustrates the training and validation loss curves over 100 epochs for MaxViT-B-22K, Swin2-B-22K, ConvNeXt-B-22K, and ViT encoders in their shallow configurations. During the first 30 epochs, the learning rate is annealed to the maximum value of $10^{-4}$ and then decreased. By this point, the ranking of models in terms of validation loss closely aligns with their final test performance. Swin2-B-22K has a significant gap between training and validation loss, yet it  achieves the lowest validation loss towards the end of training. Prov-GigaPath exhibits the highest training and validation loss throughout, and notably, its validation loss is lower than its training loss, suggesting potential underfitting. Virchow2 and ViT-H-142M demonstrate well-aligned training and validation loss curves, indicating minimal overfitting, although Virchow2 achieves better performance on the test set, as illustrated previously in Table  \ref{pannukeperformance}. 
UNI2 reaches its minimum validation loss before epoch 80, after which validation loss increases.

On CoNIC and CytoDArk0, ConvNeXt-B-22K delivers the best overall performance, followed by Swin2-B-22K and MaxViT-B-22K. UNI2 and Virchow2 are less competitive, though Virchow2 surpasses UNI2 on CytoDArk0, while UNI2 performs better on CoNIC. MaxViT-B-22K records the lowest P and DQ on CytoDArk0 but the highest mPQ+ on CoNIC.

Qualitative results for Swin2-B-22K and ConvNeXt-B-22K are presented in Fig. \ref{pannukequalit} and \ref{cytodark0equalit} for two test patches from the PanNuke and CytoDArk0 datasets, respectively. In both cases, predictions align well with the ground truth (GT). Further improvements may be possible through encoder fine-tuning, which is beyond this study's scope.

Our findings suggest that, within the scope of possibilities explored in this study, the representation learning capabilities of pre-trained ViT models can be outperformed by those of non-ViT models when applied to CS and CC in histology images. Specifically, general-purpose non-ViT models tend to achieve better results than histopathology-specific ViT models. When considering DQ, SQ, PQ, mPQ+ on PanNuke, the Swin2-B-22K exhibits a performance gap of +0.52, +0.74, +1.01, +2.76 percentage points against UNI2 (shallow), +0.21, +2.10, +1.80, +2.95 against UNI2 (mixed), +0.96, +1.14, +1.66, +3.35 against Virchow2 (shallow) and +0.46, +2.02, +2.03, +2.64 against Virchow2 (mixed). On CoNIC, ConvNeXt-B-22K  exhibits a performance gap of +5.72, +5.20, +8.21, +3.32 against UNI2 (mixed) and +6.22, +5.19, +8.56, +4.15 against Virchow2 (mixed). For DQ, SQ, PQ on CytoDArk0, ConvNeXt-B-22K exhibits a performance gap of +9.47, +7.02, +12.22 against UNI2 (mixed) and +7.16, +6.70, +10.25 against Virchow2 (mixed). On CytoDArk0, the advantage of general-purpose models is less surprising, given the differences in staining (Nissl), tissue type (brain), and cell types (neurons/glia) compared to those seen during the training of histopathology-specific foundation models. However, our results on PanNuke and CoNIC raise a main question: why does a histopathology-specific foundation model underperform compared to a general-purpose one on histopathology data? While we do not provide a definitive answer, we explore a few hypotheses in the next section.

\section{DISCUSSION}

Semantic segmentation, and instance segmentation as a particular case, inherently faces a trade-off between semantic understanding and spatial localization: global context determines \textit{what} is present, while local information identifies \textit{where} it is located \cite{long2015fully}. Feature hierarchies capture both through a nonlinear local-to-global pyramid. In convolutional encoders, feature maps progressively evolve from capturing low-level features such as edges and textures to high-level semantic representations, often resulting in the loss of fine-grained spatial details. In encoder-decoder architectures, skip connections enable the decoder to leverage this multi-scale feature representation by combining deep, coarse semantic features from the deeper layers of the encoder with shallow, fine-grained features from the shallower layers. Unlike CNNs, ViTs maintain more uniform feature representations across blocks due to their self-attention mechanism, which facilitates the early aggregation of global information and integrates contextual cues from distant regions within the image \cite{raghu2021vision}. Cells are typically small structures that do not occupy large regions of the input image. The global context, essential for diagnostic and prognostic assessments, should enhance the overall comprehension of image semantics and aid in identifying cell type compositions, as these types are often interconnected and influenced by the surrounding tissue microenvironment. However, local information is crucial for precise detection and segmentation. The inductive biases inherent in convolutional networks and Swin Transformers—such as spatial locality and hierarchical feature learning—may be critical for accurate CS and CC. In contrast, specialized foundation models with a ViT backbone rely on multiple blocks of multi-head self-attention, which may overly diffuse fine-grained information, impacting the detection and segmentation quality of small structures like cells. This may explain why ViT encoders are outperformed by non-ViT encoders, particularly ConvNeXt and the Swin Transformer V2, which combines the benefits of locality with attention mechanisms while retaining the hierarchical representation characteristic of convolutional architectures. 

This is not the sole factor that may explain our results. 
%Several factors may contribute to the observed results. 
While models pre-trained on ImageNet-22K used a supervised learning approach, others underwent unsupervised pre-training, which could influence performance differences. At the architectural level, we used 1×1 convolutions to reduce the number of channels in the feature maps extracted from the encoder (cf. Section \ref{Decoder}). Increasing the number of channels may help mitigate dimensionality loss in ViTs' latent space.
%While the models pre-trained on ImageNet-22K were trained using a supervised learning approach, the other models underwent unsupervised pre-training. 
%This distinction in training paradigms could also contribute to their relative performance differences. 
%Furthermore, at the architectural level, as described in Section \ref{Decoder}, we employed 1×1 convolutions to reduce the number of channels in the feature maps extracted from the encoder to predefined values at each level. It is possible that increasing these channel reductions could help mitigate the severity of dimensionality reduction in the latent space for ViTs. 
Additionally, ViTs process images by dividing them into patches before encoding them. The image size and patch size used during training directly affect the model’s parameterization. At inference time, images must either be rescaled to match the training resolution or require interpolation to adjust the patch embeddings and the Conv2D projection layer. This preprocessing step may impact performance, contributing to the observed differences. Further experiments are needed to determine whether rescaling would be more effective.

Our findings also highlight the crucial impact of feature depth, with ViT shallow or mixed configurations outperforming deep ones. In a Swin Transformer V2 or ConvNeXt, the choice of blocks from which to extract features is somehow obliged, since these models have four \textit{stages} that group a series of \textit{blocks}, and it is customary to take the feature maps outputted by the last block for each stage. This is the choice supported in the \texttt{timm} library. In ViTs, any block can be chosen, and this flexibility translates into an optimization problem: which combination is the best? 
%Another factor that may contribute to the underperformance of foundation models on the PanNuke dataset is the 40x magnification, whereas all foundation models except Virchow2 were trained on 20x magnification data. 

\section{CONCLUSIONS}
This study highlights the superior representation capabilities of the general-purpose Swin Transformer V2 and ConvNeXt encoders, pre-trained on ImageNet-22K, in the context of CS and CC, outperforming both general-purpose and histopathology-specific ViT-based foundation models. Applied without fine-tuning, these encoders emerge as the most robust out-of-the-box choice. The underperformance of ViT-based models may stem from their lack of inductive biases, such as locality and hierarchical representation learning, though additional factors warrant further investigation. These include differences in learning paradigms, dimensionality reduction in feature maps, and the effects of patchification and interpolation. Moreover, as only a limited number of ViT encoder configurations were explored, further optimization of block combinations for feature extraction could enhance performance. These findings emphasize the need for deeper exploration of ViT encoders while offering insights to advance CS and CC tasks.


%%%%%%%%%%%%%%%%%%%%%%%%%%%%%%%%%%%%%%%%%%%%%%%%%%%%%%%%%%%%%%%%%%%%%%%%%%%%%%%%

%%%%%%%%%%%%%%%%%%%%%%%%%%%%%%%%%%%%%%%%%%%%%%%%%%%%%%%%%%%%%%%%%%%%%%%%%%%%%%%%

%%%%%%%%%%%%%%%%%%%%%%%%%%%%%%%%%%%%%%%%%%%%%%%%%%%%%%%%%%%%%%%%%%%%%%%%%%%%%%%%

%%%%%%%%%%%%%%%%%%%%%%%%%%%%%%%%%%%%%%%%%%%%%%%%%%%%%%%%%%%%%%%%%%%%%%%%%%%%%%%%

\bibliographystyle{plain}  % Choose a bibliography style (e.g., plain, unsrt, alpha, apalike, etc.)
\bibliography{refs.bib}  % Reference the .bib file (without the .bib extension)

\end{document}